\documentclass[lettersize,journal]{IEEEtran}

\usepackage{amsmath,amsfonts}
\usepackage{algorithmic}
\usepackage{algorithm}
\usepackage{array}
\usepackage[caption=false,font=normalsize,labelfont=sf,textfont=sf]{subfig}
\usepackage{textcomp}
\usepackage{stfloats}
\usepackage{url}
\usepackage{color}
\usepackage{verbatim}
\usepackage{graphicx}
\usepackage{cite}
\usepackage{booktabs}

\usepackage{pifont}
\usepackage{algorithm}  
\usepackage{algorithmic} 

\usepackage{multirow}
\usepackage[table]{xcolor}

\newcommand{\eg}{\textit{e}.\textit{g}.}
\newcommand{\etal}{\textit{et al}.}
\usepackage{booktabs}
\newcommand{\red}[1]{\textcolor{red}{#1}}
\newcommand{\xblue}{\textcolor{black}}

\def\BibTeX{{\rm B\kern-.05em{\sc i\kern-.025em b}\kern-.08em
    T\kern-.1667em\lower.7ex\hbox{E}\kern-.125emX}}
\usepackage{balance}
\begin{document}
\title{Boosting HDR Image Reconstruction via Semantic Knowledge Transfer}
\author{
        Tao Hu,
        Longyao Wu,
        Wei Dong,
        Peng Wu,
        Jinqiu Sun,
        Xiaogang Xu,\\ 
        Qingsen Yan,
        and~Yanning Zhang,~\IEEEmembership{IEEE Fellow}
\thanks{
Qingsen Yan is with the School of Computer Science, Northwestern Polytechnical University, Xi’an 710072, China, and also with the Shenzhen Research Institute of Northwestern Polytechnical University, Shenzhen 518057, China (e-mail:qingsenyan@nwpu.edu.cn).\\
Tao Hu, Peng Wu, and Yanning Zhang are with the School of Computer Science, Northwestern Polytechnical University, Xi'an 710072, China (e-mail: taohu@mail.nwpu.edu.cn; pengwu@nwpu.edu.cn;‌ ynzhang@nwpu.edu.cn).\\
Jinqiu Sun is with the School of Astronautics, Northwestern Polytechnical University, Xi'an 710072, China.\\
Longyao Wu is with the Shaanxi University of Science \& Technology.\\
Wei Dong is with the Xi'an University of Architecture and Technology.\\
Xiaogang Xu is with the Chinese University of Hong Kong. (e-mail: xiaogangxu00@gmail.com)
}
\thanks{Corresponding author: Qingsen Yan}
}


\maketitle
\begin{abstract}
Recovering High Dynamic Range (HDR) images from multiple Standard Dynamic Range (SDR) images become challenging when the SDR images exhibit noticeable degradation and missing content. 
Leveraging scene-specific semantic priors offers a promising solution for restoring heavily degraded regions. However, these priors are typically extracted from sRGB SDR images, the domain/format gap poses a significant challenge when applying it to HDR imaging. 
To address this issue, we propose a general framework that transfers semantic knowledge derived from SDR domain via self-distillation to boost existing HDR reconstruction. Specifically, the proposed framework first introduces the Semantic Priors Guided Reconstruction Model (SPGRM), which leverages SDR image semantic knowledge to address ill-posed problems in the initial HDR reconstruction results. 
Subsequently, we leverage a self-distillation mechanism that constrains the color and content information with semantic knowledge, aligning the external outputs between the baseline and SPGRM. 
Furthermore, to transfer the semantic knowledge of the internal features, we utilize a Semantic Knowledge Alignment Module (SKAM) to fill the missing semantic contents with the complementary masks.
Extensive experiments demonstrate that our framework significantly boosts HDR imaging quality for existing methods without altering the network architecture.

\end{abstract}


\section{Introduction}

\IEEEPARstart{M}{ultiple}-exposure High Dynamic Range (HDR) imaging aims to accurately recover images with a wide luminance range from exposure-varied Standard Dynamic Range (SDR) images. 
However, SDR images are often captured in dynamic scenes, and existing deep-learning-based methods primarily focus on the aligned SDR sequences and fuse them to obtain high-quality HDR results \cite{wang2021deep}.
Nevertheless, in practical scenarios, SDR images often exhibit significant quality degradation in dark scenes due to limited light capture, while bright scenes may suffer from critical content loss caused by motion blur or occlusion, as longer exposure times required for adequate brightness increase susceptibility to object movement.
Despite advancements in recent network designs (\eg, CNNs \cite{Kalantari2017Deep,yan2019attention,BracketIRE,HDRGAN}, Transformer-based methods \cite{liu2022ghost,song2022selective,Tel_2023_ICCV,chen2023improving}), the inherently ill-posed nature of captured SDR images presents challenges in achieving substantial improvements solely through modifications to network structures.
\begin{figure}[t]
\centering
\includegraphics[width=0.9\linewidth]{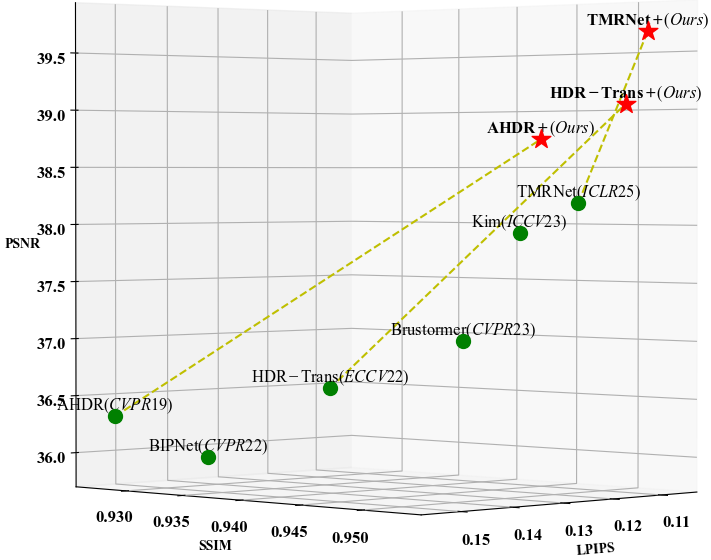} 
\caption{Performance comparison on the BracketIRE dataset, which consists of RAW format data pairs, demonstrates that our framework enables the baseline model to achieve significant improvements in \textit{PSNR} (pixel accuracy), \textit{SSIM} (detail preservation), and \textit{LPIPS} (perceptual fidelity) without altering the network architecture.}
\label{ins}
\end{figure}

Previous research \cite{wang2018recovering,zhang2024distilling,wu2023learning,liang2022semantically} has demonstrated the potential of using semantic priors to address ill-posed problems in low-level vision tasks. By integrating these priors, models can gain a deeper understanding of image content, providing explicit instructions that guide the reconstruction process. However, no work to date has explored this methodology specifically for multi-exposure HDR imaging. 
In current low-level vision tasks, the predominant approach \cite{wang2018recovering,wu2023learning,liang2022semantically} involves directly incorporating semantic priors derived from degraded images into the reconstruction network, thereby guiding the model to produce high-quality results.
However, these approaches face significant challenges when applied to HDR imaging: 
First, existing prior extraction models are primarily designed for sRGB-format SDR images. The domain gap between HDR and SDR not only complicates the generation of high-quality priors suitable for HDR imaging but also limits their effective integration into the HDR workflow. This challenge is particularly pronounced when handling RAW-format HDR datasets \cite{zou2023rawhdr}, leading to suboptimal performance in HDR tasks. 
Furthermore, HDR imaging relies on multiple SDR images with varying exposures, where semantic information can vary significantly across images, leading to incomplete or conflicting priors from each input.

The aforementioned issues motivate us to leverage semantic priors from another perspective. The motivation for our work is threefold. First, given the variations and conflicts in semantics within each SDR image, we exploit fused HDR images to extract high-quality semantic priors. To this end, we employ a pre-trained semantic segmentation network as a Semantic Knowledge Bank (SKB), which consists of semantic features and segmentation maps. Second, considering the domain gap between HDR and SDR, we perform prior embedding in the SDR domain to enhance the quality of the initial HDR results, fully harnessing the potential of semantic knowledge. Third, inspired by the semantic consistency across different formats and domains, we distill semantic knowledge from the SDR domain to boost the performance of existing HDR models.

Based on the above analysis, we propose a general framework that effectively leverages the advanced capabilities of semantic priors for HDR imaging. As shown in Fig.~\ref{fw}, our proposed framework comprises the Original Reconstruction Model (ORM), the Semantic Priors Guided Reconstruction Model (SPGRM), and the Semantic Knowledge Transfer Scheme (SKTS). The ORM can be any existing HDR imaging method, tasked with generating initial HDR results. The SPGRM utilizes intermediate semantic features derived from the SKB to refine the reconstructed image. Notably, the ORM operates in the HDR domain, whereas the SPGRM functions in the SDR domain. 
Subsequently, the SKTS exploits semantic consistency across formats/domains, using knowledge distillation \cite{zhang2021self} to ensure the ORM aligns with the semantic knowledge embodied in the SPGRM. Unlike existing distillation methods that solely distill priors in the level of image content, 
 we also introduce a Semantic-Guided Histogram Loss to enhance color consistency across distinct object instances.
Furthermore, inspired by MAE’s masked reconstruction \cite{he2022masked}, we introduce a Semantic Knowledge Alignment Module to assist SKTS, ensuring consistency in the semantic feature space during the distillation process.  As illustrated in Fig.~\ref{ins}, our framework significantly enhances the performance of all baseline models. The main contributions can be summarized as follows:
\begin{itemize}
    \item {We propose a general semantic knowledge transfer framework that leverages self-distillation technology to boost the performance of existing multi-exposure HDR reconstruction models, without modifying the original network architecture.}
    \item {We introduce a Semantic-Guided Histogram Loss and a Semantic Knowledge Alignment Module to enable effective semantic knowledge distillation at both the feature and output levels.}
    \item {We conduct experiments on several RAW/sRGB format datasets, demonstrating that our method can significantly improve the HDR imaging quality of existing methods.''}
\end{itemize}

\section{Related Work}
\subsection{Multi-exposure HDR Imaging}
{Multi-exposure HDR imaging has been a pivotal area of research in computational photography, aiming to reconstruct high-quality HDR images from a set of SDR images captured at different exposure levels \cite{Banterle:2017}. Traditional methods in this domain have primarily relied on pre-processing techniques such as motion rejection and motion registration \cite{Heo2011ACCV,yan2017high,Ward2012,Tomaszewska07,sen2012robust,Hu2013deghosting}. These techniques are designed to align the SDR images and mitigate artifacts caused by motion between exposures. However, their effectiveness diminishes in scenarios involving large-scale object movements or complex scene dynamics, as they heavily depend on the accuracy of preprocessing steps.
In recent years, deep learning-based methods have emerged as a promising alternative, researchers have explored a variety of deep neural network (DNN) architectures and techniques. For instance, attention mechanisms \cite{yan2023unified,chen2022attention,yan2019attention,10758406} have been employed to selectively focus on relevant image regions, enhancing alignment and fusion accuracy. Similarly, transformer-based models \cite{song2022selective,Tel_2023_ICCV,chen2023improving,liu2023joint} leverage global contextual understanding to improve robustness in HDR reconstruction. Optical flow-based approaches \cite{Kalantari2017Deep,xu2024hdrflow,Kong_2024_ECCV} estimate motion between frames to compensate for misalignment, while generative adversarial networks (GANs) \cite{niu2021hdr,9826814} and diffusion models \cite{hu2024generating,yan2024dynamic,10288540} have been adapted to synthesize realistic HDR content from imperfect inputs. Additional innovative frameworks, such as \cite{zhang2023lookup,fang2024glgnet,li2025afunet,ni2025semantic,xu2025adaptiveae}, further expand the scope of DNN applications in this domain. AFUNet\cite{li2025afunet} adopts a deep-unfolding formulation with cross-iterative coupling between alignment and fusion, which strengthens motion compensation and detail aggregation in dynamic scenes. SMHDR\cite{ni2025semantic} introduces semantic masking together with a curriculum-learning schedule that progressively suppresses saturated or misaligned regions, thereby improving reconstruction robustness. 
FlexHDR \cite{9881970} models alignment and exposure uncertainties to produce high quality HDR results.
AdaptiveAE\cite{xu2025adaptiveae} proposes an adaptive exposure strategy for dynamic scenes; although orthogonal to network architecture, it complements reconstruction by reducing saturation and ghosting at capture time.}

Despite these advancements, existing multi-exposure HDR imaging methods primarily focus on addressing artifacts caused by motion. They often overlook the inherently ill-posed nature of captured SDR images in real-world scenarios, where images may suffer from noticeable degradations or information loss due to factors like noise, blur, or saturation. Consequently, the reconstruction quality of HDR images is significantly reduced when the input SDR images are of suboptimal quality. This limitation underscores the need for methods that can robustly handle degraded inputs while maintaining high reconstruction fidelity.

\begin{figure*}[t]
\centering
\includegraphics[width=0.98\textwidth]{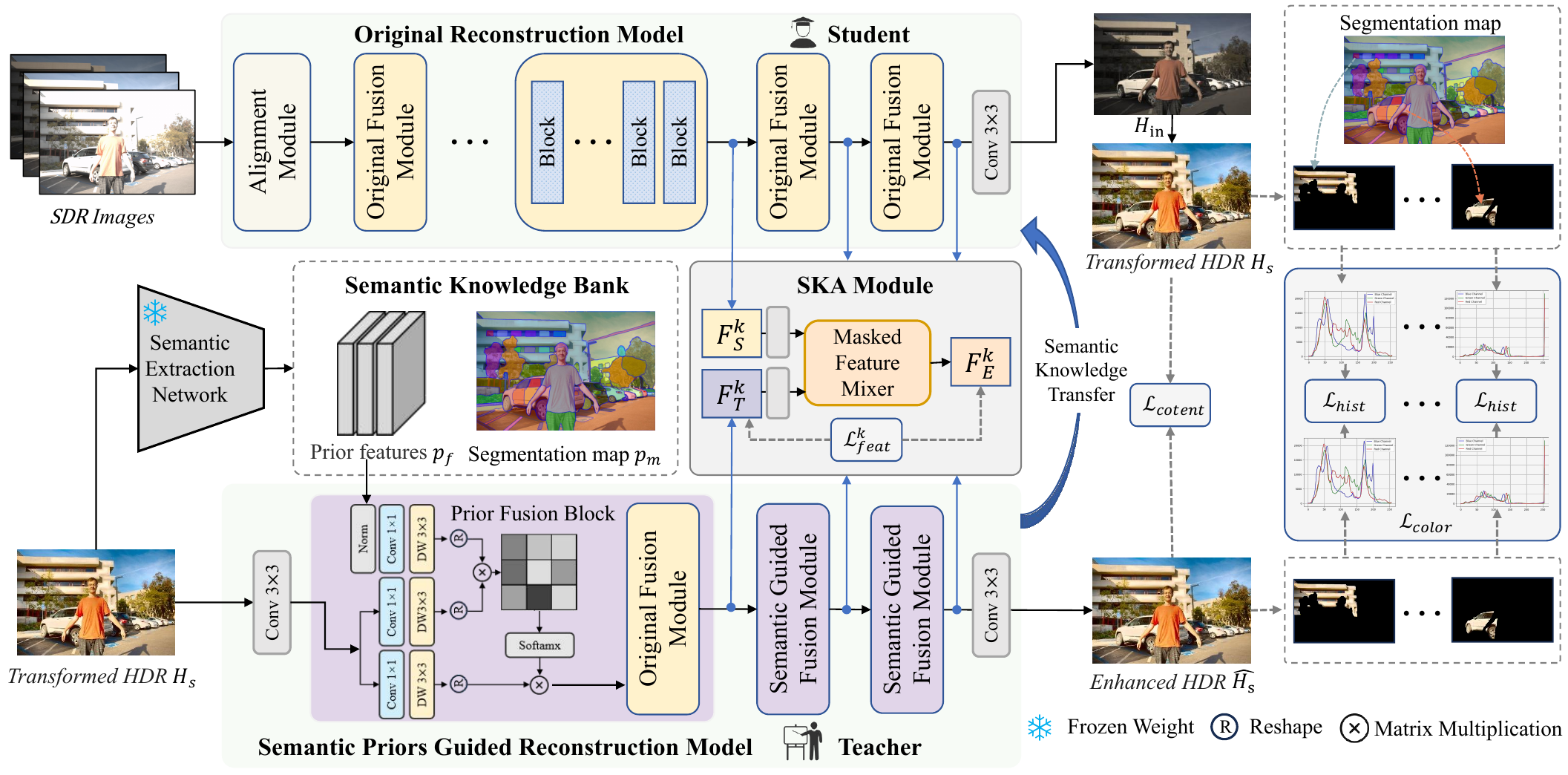} 
\caption{The Proposed Semantic Knowledge Transfer Framework. During training, the SPGRM enhances ORM's initial HDR results by integrating semantic priors. Meanwhile, the Semantic Knowledge Transfer scheme boosts the performance of the ORM through model/feature-level self-distillation. During inference, only the optimized ORM is deployed, ensuring high efficiency.}
\label{fw}
\end{figure*}

\subsection{Semantic Priors for Image Reconstruction.}
In the broader context of image reconstruction, particularly in low-level visual tasks such as super-resolution, denoising, and deblurring, the use of semantic priors has emerged as a powerful strategy to address ill-posed problems. Semantic priors provide high-level contextual information that can guide the reconstruction process, enabling models to produce more accurate and perceptually pleasing results.  
One prominent approach is the use of semantic-loss-guided methods \cite{aakerberg2022semantic,liu2018when,zheng2022semantic,zhang2025high}, which incorporate semantic segmentation loss into the training process. This strategy explicitly enforces the model's ability to capture deep semantic features, thereby enhancing its performance in reconstructing images with complex structures. For example, Wu \etal \cite{wu2023learning} and Liang \etal \cite{liang2022semantically} utilize segmentation maps as priors to design tailored loss functions, which boost the overall performance of the model by aligning the reconstruction with semantic boundaries.  
Another category of methods involves the direct integration of semantic information into the model's architecture. Segmentation-map-guided methods \cite{wang2018recovering,zhang2024distilling,fan2020integrating,zheng2022semantic} and intermediate-feature-guided methods \cite{wu2023learning,li2022close,li2020blind} embed semantic priors, such as segmentation maps or intermediate features from pre-trained networks, to condition the reconstruction process. SFTGAN \cite{wang2018recovering}, for instance, was one of the earliest methods to use semantic segmentation maps as priors to guide models in low-level visual tasks, demonstrating significant improvements in texture and detail recovery. 
{Semantic priors have also been applied in HDR contexts, such as inverse tone mapping. For example, Goswami \etal \cite{goswami2024semantic} use semantic segmentation and an ordered graph to guide diffusion-based inpainting for reconstructing clipped regions in single SDR images, lifting them to HDR.}

Despite their success in the SDR domain, these semantic-prior-guided methods face challenges when extended to HDR imaging. A significant limitation is the domain gap between SDR and HDR images, particularly since HDR images are typically stored in RAW format, which differs substantially from the processed SDR images used in most existing methods. This discrepancy necessitates careful adaptation of semantic priors to ensure they are compatible with the unique characteristics of HDR data. Bridging this gap remains a critical challenge for effectively applying semantic prior-guided approaches to HDR contexts.

\section{Method}
\subsection{Overview of Proposed Framework}
As shown in Fig. \ref{fw}, our framework primarily consists of the Original Reconstruction Model (ORM) and the Semantic Priors Guided Reconstruction Model (SPGRM), with knowledge self-distillation performed through the Semantic Knowledge Transfer Scheme (SKTS). Given a series of SDR images of a dynamic scene $\left\{L_1, L_2, \ldots, L_n\right\}$, the ORM utilizes existing HDR reconstruction methods to produce the initial HDR images $H_{in}$. {These initial HDR images are then converted into tonemapped sRGB domain results $H_{s}$ through a domain transfer operation:
\begin{equation}
H_{in}=f_{org}(L_1, L_2, \ldots, L_n ; \theta_o), 
H_{s} = \mathcal{T}(H_{in}),
\end{equation}
where $\mathcal{T}(\cdot)$ represents the domain transfer operation, which for sRGB format data includes a $\mu$-law function, and for RAW format data additionally includes a differentiable demosaicing operation; $f_{org}$ denotes the ORM network, and $\theta_o$ is the network parameter. }
To extract semantic priors, the subsequent process can be represented as:
\begin{equation}
p_f,p_m={f}_{\text {seg}}\left(H_{s} ; \theta_s\right),
\end{equation}
where $p$ is the semantic prior, including segmentation maps $p_m$ and intermediate features $p_f$ with multi-scale dimensions. ${f}_{\text {seg}}$ and $\theta_s$ represent the pre-trained segmentation network and its frozen parameters. Then, $p_f$ and $H_{s}$ are used as inputs to the SPGRM to obtain the enhanced reconstruction result $\widehat{H_s}$:
\begin{equation}
\widehat{H_s}=f_{\text {e}}\left(H_{s}, p_f ; \theta_e\right),
\end{equation}
where $f_{\text {e}}$ and $\theta_e$ are the network and corresponding parameters of the SPGRM. 

During the training stage, we use the SKTS to embed the semantic knowledge from $f_{\text {e}}$ into $f_{\text {org}}$  through the self-distillation manner.
$\theta_e$ and $\theta_o$ will be updated by minimizing the objective function while $\theta_s$ is fixed. During the inference stage, only $f_{\text {org}}$ participates in the inference, without requiring $f_{\text {e}}$ and $f_{\text {seg}}$ to be involved.

\subsection{Semantic Priors Guided Reconstruction Model}
To avoid semantic variations and conflicts in each SDR image, we utilize ORM's result to extract semantic priors. However, when leveraging these priors to guide the model in reconstructing high-quality images, another challenge that requires careful attention is the disparity between the prior and feature sources. To address this, we implement the SPGRM in the sRGB SDR domain, producing sRGB-tonemapped HDR results. 
Notably, the ORM and the SPGRM are trained simultaneously, and the transformation operation does not compress any of the original HDR content or impede gradient backpropagation.

As illustrated in Fig. \ref{fw}, existing HDR imaging models typically follow an ``align-then-fuse" paradigm, implicitly aligning SDR images before fusing them to produce HDR results. Given that the SPGRM takes the ORM's results as input, it solely consists of three Semantic Guided Fusion Modules and excludes the SDR alignment module.
{In our framework, we employ FastSAM \cite{zhao2023fastsegment} as the SKB owing to its outstanding performance and ease of use. Compared to SAM \cite{kirillov2023segment}, FastSAM achieves inference speeds tens of times faster while maintaining robust semantic prior extraction capabilities.} We leverage the output features before the representation head to get four multi-scale semantic features $p_f^i\in \mathbb{R}^{  {C_i} \times {\frac{H}{2^{4 - i}}} \times {\frac{W}{2^{4 - i}}}} (i = 0,1,2,3)$. Here, $C_i$ represents the channel dimension, and $H$ and $W$ denote the height and width of the input image, respectively. 
These $p_f^i$ features are processed through a Feature Pyramid Network (FPN) \cite{lin2017feature} for multi-level feature interaction, yielding $p_f$, which is adjusted to match the channel dimensions of SPGRM's intermediate features. To incorporate the semantic priors $p_f$, we add a Prior Fusion Block to generate the refined feature map $\bar{F_k^T}$ within each fusion module.
Specifically, as depicted in Fig. \ref{fw}, the Prior Fusion Block takes two inputs: an intermediate feature $F_k^T$ and a prior feature $p_f$. We project $p_f$ into a query vector $\mathbf{Q} = W_d^Q W_c^Q p_f$ using point-wise $1\times 1$ convolution and depth-wise $3\times 3$ convolution with weights $W_d^Q$ and $W_c^Q$, respectively. Similarly, $F_k^T$ is transformed into Key $\mathbf{K}$ and Value $\mathbf{V}$ through analogous operations.
Then, we adopt a transposed-attention mechanism \cite{zamir2022restormer} to compute the attention map. Hence, the process can be described as follows:
\begin{equation}
{\bar{F_{k}^T}}=W_c {{V}} \cdot \operatorname{Softmax}({{K}} \cdot {{Q}} / \gamma)+{F}_{k}^{T},
\end{equation}
where $\gamma$ is a trainable parameter. 
\subsection{Semantic Knowledge Transfer Scheme}
During the initial training phase, the output of SPGRM produces a high-quality tone-mapped HDR image $\widehat{H_s}$, which incorporates semantic priors and exhibits improved image quality compared to the initial HDR image ${H_{in}}$. To bridge this performance gap and enable the ORM to match or approximate the SPGRM's capabilities, we introduce a semantic knowledge transfer scheme, which utilizes SPGRM as a teacher signal to guide the training of the ORM. The key aspects of the scheme are as follows:

\textbf{Preliminary alignment.}
When utilizing the SPGRM as the teacher model, a critical challenge arises from disparities in the outputs, including domain mismatches (\eg, HDR \textit{vs.} SDR) and representation differences (\eg, RAW \textit{vs.} sRGB), between the SPGRM and ORM. 
To address this challenge, we propose a simple yet effective approach: before performing self-distillation, we introduce a domain transfer operator $\mathcal{T}(\cdot)$ to align the outputs of the two models across domains and representations. Provided the transfer function is reversible and differentiable, optimizing the network in the transformed domain implicitly boosts the student model’s performance. Notably, the transformation operation does not compress any of the original HDR content or impede gradient backpropagation.

\textbf{Semantic Knowledge distillation.}
Unlike previous work that focuses solely on knowledge distillation at the content level, we recognize the importance of color alongside content in HDR reconstruction tasks. Therefore, our scheme executes knowledge transfer at both the content and color levels. For the content level, we distill the semantic priors from the SPGRM to the ORM by minimizing the $\mathcal{L}_1$ loss in pixel space and the perceptual loss in the feature representation space:
\begin{equation}
\scriptstyle
\mathcal{L}_{content}=\left\|\mathcal{T}(H_{in})-\widehat{H_s}\right\|_1 
+\lambda\left\|\phi_{i, j}(\mathcal{T}(H_{in}))-\phi_{i, j}(\widehat{H_s})\right\|_1,
\end{equation}
where $\mathcal{T}(\cdot)$ denotes the domain transfer operator; $\widehat{H_s}$ is the output of the SPGRM, and $H_{in}$ is the output of the ORM. $\phi_{i,j}(\cdot)$ signifies the $j$-th convolutional feature extracted from the VGG19 network after the $i$-th maxpooling operation, and $\lambda=1e-2$ is a balance hyperparameter. 

At the color level, unlike conventional color loss functions \cite{zhang2022deep} that focus on global statistical properties, our approach tackles the significant brightness variations across different regions in HDR images by enforcing constraints on distinct object instances.
{Given the discrete nature of color histograms, we propose a Semantic-Guided Histogram Loss that leverages Kernel Density Estimation (KDE) \cite{zambom2013review} to compute a differentiable histogram.
Specifically, we first define the bin centers of the histogram. For a total of \( N \) bins, the center of the $i$-th bin $c_i$ is computed as:
\begin{equation}
c_i = \min + \frac{\max - \min}{N} (i + 0.5), \quad i \in \{0, \dots, N-1\},
\end{equation}
where \( \max \) and \( \min \) represent the upper and lower bounds of the pixel values, respectively.
Let \( x_{c,j} \in \mathbb{R}^{3 \times H \times W} \) denote the \( j \)-th pixel value in the \( c \)-th channel of the input image. The Gaussian kernel value between \( x_{c,j} \) and the \( i \)-th bin center \( c_i \) is computed as:
\begin{equation}
K(x_{c,j}, c_i) = \exp\left(-\frac{(x_{c,j} - c_i)^2}{\sigma^2}\right),
\end{equation}
where \( \sigma \)=400 is the bandwidth parameter. The approximate histogram value $H_{c,i}$ of the $c$-th channel is:
\begin{equation}
H_{c,i} = \sum_{j=1}^{H \times W} K(x_{c,j}, c_i).
\end{equation}
To obtain a proper probability distribution, we normalize the histogram by dividing each $H_{c,i}$ by the sum of $H_c$ for numerical stability.
Then, the mean squared error (MSE) between the $\mathrm{CDF}_{c, k}=\sum_{i=0}^k H_{c, i}, k \in\{0, \ldots, N-1\}$ of the two histograms, is used as the histogram loss:
\begin{equation}
\mathcal{L}_{\text {hist }}=\frac{1}{C} \sum_{c=1}^C \sum_{k=0}^{N-1}\left(\mathrm{CDF}_{c, k}^{\mathrm{S}}-\mathrm{CDF}_{c, k}^{\mathrm{T}}\right)^2,
\label{hist}
\end{equation}
where $\mathrm{CDF}_{c, k}^{\mathrm{S}}$ and $\mathrm{CDF}_{c, k}^{\mathrm{T}}$ represent the CDFs derived from the color histograms of the output of the student and teacher model, respectively.
Building on Eq.~\eqref{hist}, the proposed $\mathcal{L}_{\text{color}}$ is formulated as:
\begin{equation}
\mathcal{L}_{\text{color}} = \frac{1}{N} \sum_{i=0}^{N-1} \text{Hist} \left( p^i_m \odot \mathcal{T}(H_{\text{in}}), p^i_m \odot \widehat{H_{\text{s}}} \right),
\end{equation}
where \( \odot \) denotes element-wise multiplication, \( p^i_m \) is the \( i \)-th channel of the binary segmentation maps derived from SKB, and \( \text{Hist}(\cdot) \) is the histogram loss computed over each region.}

\begin{figure}[t]
\centering
\includegraphics[width=1\linewidth]{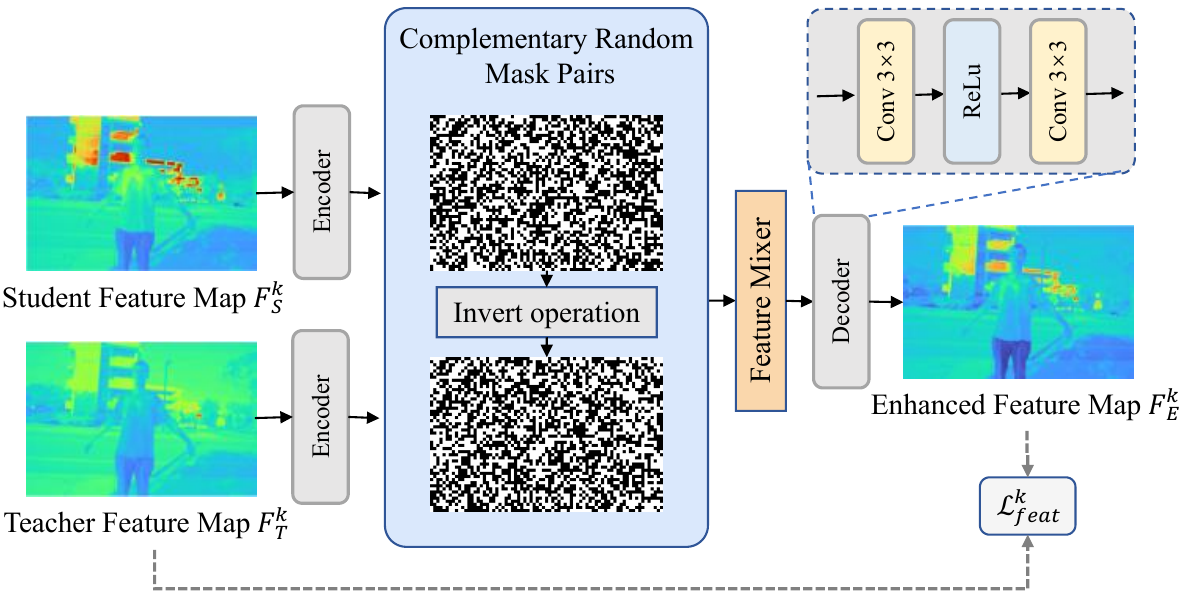} 
\caption{{The process of Semantic Knowledge Alignment Module, which conducts semantic knowledge transfer at the feature level.}}
\label{SKAM}
\end{figure}

\subsection{Semantic Knowledge Alignment Module}
Beyond model-level distillation, semantic transfer at the feature-level is equally essential. However, since the ORM and SPGRM operate in different domains or formats, directly aligning their features can impair model performance. To address this issue, we introduce the Semantic Knowledge Alignment Module (SKAM), which progressively enhances semantic knowledge consistency at the feature-level within a unified latent space.

The motivation behind SKAM is twofold. First, since different domains/formats share the same semantic information, knowledge transfer of feature-level can be achieved within a unified latent space. Second, inspired by the masked reconstruction \cite{li2024knowledge}, we aim to perform feature-level distillation in a more gradual manner, rather than directly minimizing feature distances. Specifically, as shown in Fig. \ref{SKAM}, {SKAM is composed of two encoders, one masked feature mixer, and one decoder. The encoders and the decoder possess the same architectural design, each consisting of two convolutional layers.} At the \( k \)-th alignment stage, feature maps from the ORM (\( F^k_S \)) and SPGRM (\( F^k_T \)) are processed by their respective encoders, mapping them into a unified latent space to produce latent representations \( \hat{F^k_S} \) and \( \hat{F^k_T} \). {These representations are then fused using a set of two randomly generated binary complementary masks, with a 0.5 probability for each value in the masks.} The decoder subsequently transforms the combined features into an enhanced feature \( F^k_E \) in the same space as \( F^k_T \), expressed as:
\begin{equation}
F_E^k = \operatorname{Decoder}\left( \hat{\mathbf{F}_S^k} \odot \left( \mathbf{1} - \mathbf{I}^M \right) + \hat{\mathbf{F}_T^k} \odot \mathbf{I}^M \right),
\end{equation}
where \( \mathbf{I}^M \in \{0,1\}^{C \times H \times W} \) denotes a randomly generated binary mask, and \( \odot \) represents element-wise matrix multiplication. To minimize the semantic knowledge representation gaps between the ORM and SPGRM at the feature-level, the knowledge alignment loss is computed as:
\begin{equation}
\mathcal{L}_{feat}^{k} = \operatorname{MSE}\left( F^k_E, F^k_T \right),
\end{equation}
where \(\operatorname{MSE}\) refers to the mean squared error.

\begin{figure}[t]
\centering
\includegraphics[width=1\linewidth]{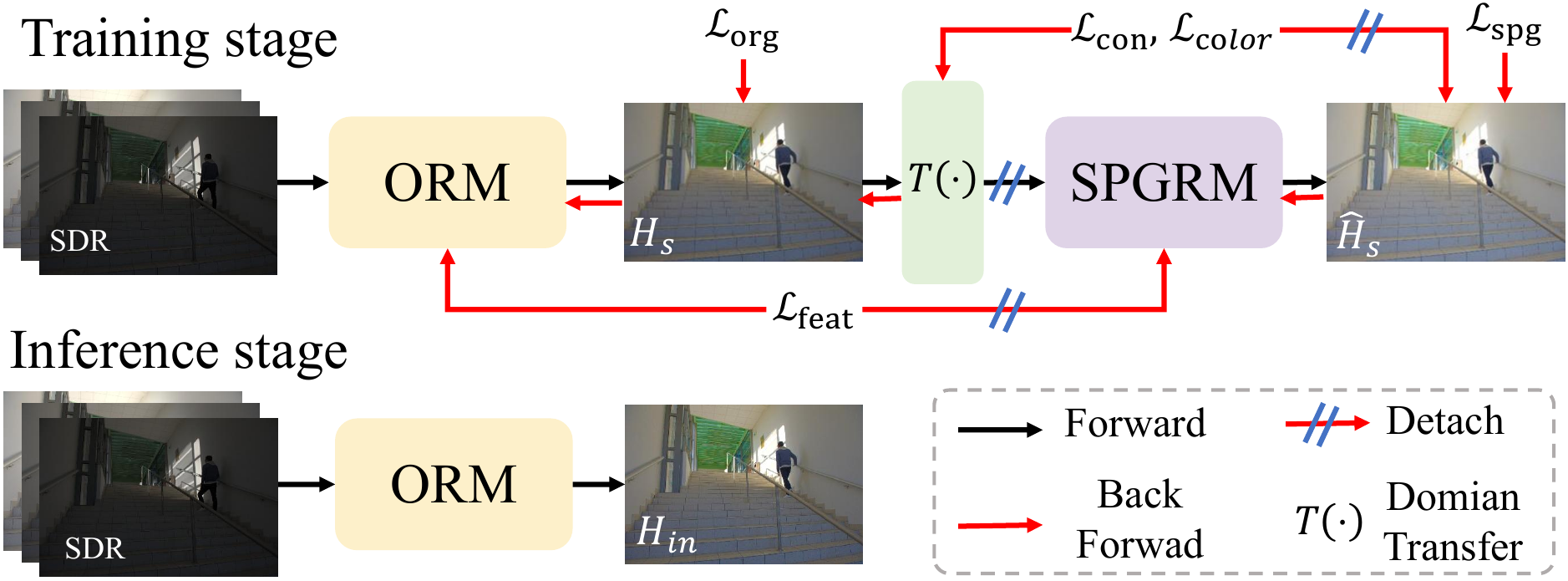} 
\caption{{The process of the training and inference stages.}}
\label{train_inf}
\end{figure}
\subsection{Training and Inference}
As shown in Fig. \ref{train_inf}, in the training stage, the SPGRM is exclusively supervised using a domain-transferred ground truth and optimized through an $\mathcal{L}_1$ loss, denoted as \( L_{\text{spg}} \):
\begin{equation}
\mathcal{L}_{spg}=\left\|\widehat{H_s}-\mathcal{T}(GT)\right\|_1,
\end{equation}
Meanwhile, the ORM is supervised using its original reconstruction loss functions, denoted as $\mathcal{L}_{org}$. The overall objective function for the ORM, denoted as:
\begin{equation}
\mathcal{L}=\mathcal{L}_{org}+\lambda_1 \mathcal{L}_{content}+\lambda_2 \mathcal{L}_{color}+\mathcal{L}_{{feat}},
\end{equation}
where $\lambda_1$ and $\lambda_2$ are hyperparameters empirically set to 1e-3 and 100. 
{To tackle the variability in the number of segmentation instances output by SAM-based models, we fix the number of output instance masks to a constant value $K=50$ for every training sample. Notably, only non-zero (valid) instance masks are involved in the calculation of $\mathcal{L}_{color}$.}
During the inference stage, only ORM participates in the inference, without requiring SPGRM.

\begin{table*}[h]
\centering
\caption{\xblue{Quantitative comparison with state-of-the-art methods on the Tel \etal's testing samples \cite{Tel_2023_ICCV}. The improvements brought by our framework to the baseline models are highlighted in red, with the SSIM improvement displayed as 100$\times$ the actual value.}}
\label{compare_1}
\scalebox{1}{
\begin{tabular}{cccccccc}
\toprule[1.2pt]
\multirow{2}{*}{Methods} & \multicolumn{7}{c}{Tel~\etal's~dataset \cite{Tel_2023_ICCV}} \\ \cline{2-8}
& PSNR-$\mu$ $\uparrow$ & SSIM-$\mu$ $\uparrow$ & PSNR-L $\uparrow$ & SSIM-L $\uparrow$ & PSNR-PU $\uparrow$ & SSIM-PU $\uparrow$ & HDR-VDP2 $\uparrow$ \\ \midrule
NHDRR \cite{yan2020deep}      & 36.68 & 0.9590 & 39.61 & 0.9853 & 37.06 & 0.9777 & 67.52 \\
DHDR \cite{Kalantari2017Deep} & 40.05 & 0.9794 & 43.37 & 0.9924 & 40.46 & 0.9889 & 68.08 \\
HDRGAN \cite{niu2021hdr}      & 41.87 & 0.9832 & 45.23 & 0.9939 & 42.13 & 0.9908 & 70.14 \\
DiffHDR \cite{10288540}       & 42.19 & 0.9843 & 45.66 & 0.9945 & 42.51 & 0.9919 & 71.52 \\
SMHDR \cite{ni2025semantic}   & 43.41 & 0.9883 & 47.46 & 0.9952 & 43.79 & 0.9941 & 72.76 \\
AFUNet \cite{li2025afunet}    & 43.31 & 0.9876 & 47.83 & 0.9959 & 43.72 & 0.9945 & 73.22 \\
SCTNet \cite{Tel_2023_ICCV}   & 42.48 & 0.9849 & 47.46 & 0.9952 & 42.79 & 0.9924 & 71.99 \\ \midrule \midrule
AHDR \cite{yan2019attention}  & 41.92 & 0.9836 & 45.20 & 0.9942 & 42.29 & 0.9919 & 70.19 \\ \hline
\multirow{2}{*}{AHDR+}        & 43.18 & 0.9874 & 46.97 & 0.9955 & 43.61 & 0.9937 & 72.34 \\
                              & \red{+1.26} & \red{+0.38} & \red{+1.77} & \red{+0.13} & \red{+1.32} & \red{+0.18} & \red{+2.15} \\ \midrule \midrule
HDR-Trans \cite{liu2022ghost} & 42.32 & 0.9843 & 46.38 & 0.9948 & 42.64 & 0.9920 & 70.99 \\ \hline
\multirow{2}{*}{HDR-Trans+}   & 43.38 & 0.9879 & 47.85 & 0.9959 & 43.84 & 0.9943 & 72.88 \\
                              & \red{+1.06} & \red{+0.36} & \red{+1.47} & \red{+0.11} & \red{+1.20} & \red{+0.23} & \red{+1.89} \\ \midrule \midrule
TMRNet \cite{BracketIRE}      & 42.72 & 0.9867 & 45.98 & 0.9953 & 43.23 & 0.9936 & 72.15 \\ \hline
\multirow{2}{*}{TMRNet+}      & 43.15 & 0.9874 & 47.19 & 0.9961 & 43.47 & 0.9939 & 72.22 \\
                              & \red{+0.43} & \red{+0.07} & \red{+1.21} & \red{+0.08} & \red{+0.24} & \red{+0.03} & \red{+0.07} \\ 
\toprule[1.2pt]
\end{tabular}}
\end{table*}

\begin{table*}[th]
\centering
\caption{\xblue{Quantitative comparison with state-of-the-art methods on the Kalantari \etal's testing samples \cite{Kalantari2017Deep}. The improvements brought by our framework to the baseline models are highlighted in red, with the SSIM improvement displayed as 100$\times$ the actual value.}}
\label{compare_2}
\scalebox{1}{
\begin{tabular}{cccccccc}
\toprule[1.2pt]
\multirow{2}{*}{Methods} & \multicolumn{7}{c}{Kalantari~\etal's~dataset \cite{Kalantari2017Deep}} \\ \cline{2-8}
& PSNR-$\mu$ $\uparrow$ & SSIM-$\mu$ $\uparrow$ & PSNR-L $\uparrow$ & SSIM-L $\uparrow$ & PSNR-PU $\uparrow$ & SSIM-PU $\uparrow$ & HDR-VDP2 $\uparrow$ \\ \midrule
NHDRR \cite{yan2020deep}      & 42.41 & 0.9877 & 41.43 & 0.9857 & 42.97 & 0.9855 & 66.69 \\
DHDR \cite{Kalantari2017Deep} & 42.67 & 0.9888 & 41.23 & 0.9846 & 41.83 & 0.9832 & 70.58 \\
HDRGAN \cite{niu2021hdr}      & 43.92 & 0.9905 & 41.57 & 0.9865 & 44.03 & 0.9851 & 71.13 \\
DiffHDR \cite{10288540}       & 44.11 & 0.9911 & 41.73 & 0.9885 & 43.81 & 0.9914 & 71.19 \\
HyHDR \cite{yan2023unified}   & 44.64 & 0.9915 & 42.47 & 0.9894 & 45.25 & 0.9928 & 71.96 \\
SMHDR  \cite{ni2025semantic}  & 44.69 & 0.9920 & 42.61 & 0.9894 & 45.30 & 0.9932 & 71.94 \\ 
AFUNet \cite{li2025afunet}    & 44.91 & 0.9923 & 42.59 & 0.9906 & 45.57 & 0.9935 & 71.98 \\
SCTNet \cite{Tel_2023_ICCV}   & 44.43 & 0.9918 & 42.21 & 0.9891 & 44.27 & 0.9927 & 71.81 \\
LF-Diff \cite{hu2024generating} & 44.76 & 0.9919 & 42.59 & 0.9906 & 45.28 & 0.9930 & 72.11 \\ \midrule \midrule
AHDR \cite{yan2019attention}  & 43.62 & 0.9900 & 41.03 & 0.9862 & 42.93 & 0.9849 & 71.08 \\ \hline
\multirow{2}{*}{AHDR+}        & 44.24 & 0.9910 & 41.54 & 0.9880 & 44.99 & 0.9925 & 71.24 \\
                              & \red{+0.62} & \red{+0.10} & \red{+0.51} & \red{+0.18} & \red{+2.06} & \red{+0.76} & \red{+0.16} \\ \midrule \midrule
HDR-Trans \cite{liu2022ghost} & 44.24 & 0.9916 & 42.17 & 0.9890 & 44.23 & 0.9924 & 70.22 \\ \hline
\multirow{2}{*}{HDR-Trans+}   & 44.68 & 0.9918 & 42.39 & 0.9897 & 45.27 & 0.9928 & 71.94 \\
                              & \red{+0.44} & \red{+0.02} & \red{+0.22} & \red{+0.07} & \red{+1.04} & \red{+0.04} & \red{+1.72} \\ \midrule \midrule
TMRNet \cite{BracketIRE}      & 43.92 & 0.9907 & 42.09 & 0.9891 & 44.41 & 0.9918 & 71.45 \\ \hline
\multirow{2}{*}{TMRNet+}      & 44.49 & 0.9915 & 42.39 & 0.9895 & 45.11 & 0.9928 & 71.79 \\
                              & \red{+0.57} & \red{+0.08} & \red{+0.30} & \red{+0.04} & \red{+0.70} & \red{+0.10} & \red{+0.34} \\ 
\toprule[1.2pt]
\end{tabular}}
\end{table*}
\section{Experiments}
\subsection{Experimental Settings}

\noindent\textbf{Datasets.}
{We evaluate the proposed framework on several datasets spanning diverse scenes, including Kalantari’s dataset \cite{Kalantari2017Deep}, Tel \etal’s dataset \cite{Tel_2023_ICCV}, and Zhang \etal’s dataset \cite{BracketIRE}. These datasets cover a wide range of scene conditions, encompassing variations in lighting, object motion, and camera motion.
Kalantari \etal’s dataset is a real-captured dataset consisting of 74 training samples and 15 testing samples, focusing primarily on daytime dynamic scenes. Zhang \etal’s dataset, by contrast, focuses on low-light environments and includes two subsets: BracketIRE and BracketIRE+. BracketIRE contains SDR image sequences with various degradations, while BracketIRE+ extends this by incorporating an additional 4× super-resolution task. Each subset comprises 1,045 pairs for training and 290 pairs for testing.
Tel \etal’s dataset offers a more balanced distribution of scenes and motions, with samples collected under diverse lighting conditions and encompassing various motion types in equal proportions; it includes 108 training samples and 36 testing samples. 
To further assess the model’s ability to generalize, we also make use of Sen \etal’s dataset \cite{sen2012robust} and Tursun \etal’s dataset \cite{Tursun2016data}; these datasets are exclusively used for qualitative evaluation, given that they do not provide corresponding ground truths.
Notably, Kalantari’s and Tel’s datasets are in sRGB format, with three SDR images per scene as input. In comparison, Zhang’s dataset uses RAW format, with five SDR images per scene as input. }
\begin{figure*}[t]
\centering
\includegraphics[width=1\textwidth]{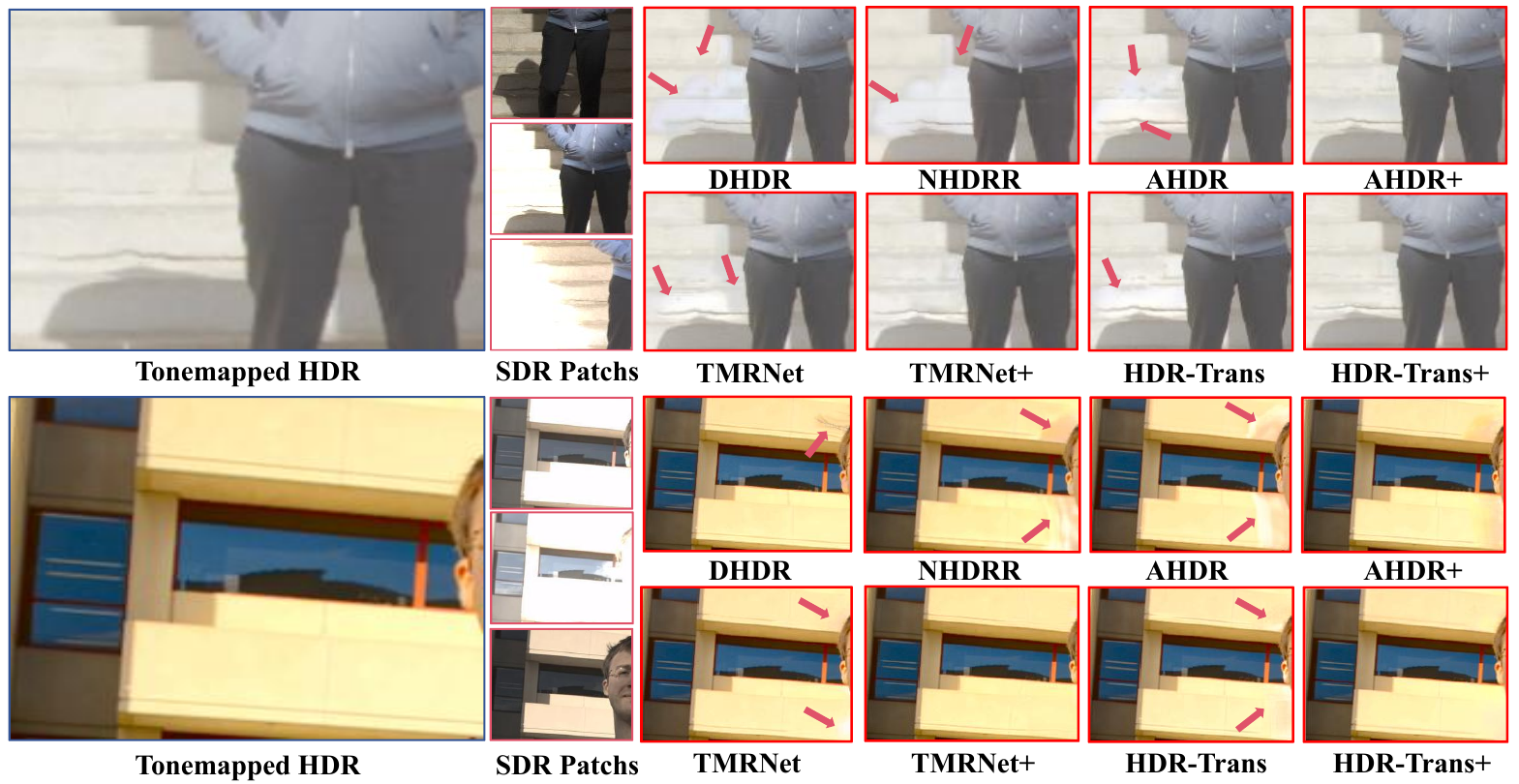} 
\caption{{Visual comparison of baseline methods with and without our framework on Tel and Kalantari’s datasets.}}
\label{compareA}
\end{figure*}

\noindent\textbf{Evaluation Metrics.}
{For Zhang \etal’s dataset in RAW format, we follow the evaluation protocol established in the original BracketIRE dataset \cite{BracketIRE}. Specifically, RAW images are first converted to linear RGB using a defined ISP pipeline, then tone-mapped to 16-bit SDR for computing PSNR, SSIM, and LPIPS \cite{zhang2018unreasonable}. All metrics are evaluated against the corresponding SDR ground truth.
For Kalantari \etal ’s and Tel \etal’s datasets in sRGB format, we adopt the same metrics as prior work \cite{yan2019attention, liu2022ghost, Tel_2023_ICCV}: PSNR-L, PSNR-$\mu$, SSIM-L, SSIM-$\mu$, and HDR-VDP2 \cite{Mantiuk2011HDR} to enable fair and intuitive comparison. The subscripts L and $\mu$ denote evaluation in the linear and $\mu$-law tone-mapped domains, respectively. 
\xblue{We calculate HDR-VDP2 using linear input images with the ``rgb-native'' color encoding configuration. The data is treated as display-referred and scaled to a peak luminance of $L_{peak} = 1000$ nits, with a viewing angle of $30^\circ$ and a viewing distance of 0.55 meters.}
In addition, we extend our evaluation suite with PSNR-PU and SSIM-PU, computed using the official PU21 implementation \cite{azimi2021pu21}, which are perceptually optimized metrics specifically designed for HDR quality assessment.}

\noindent\textbf{Compared Methods.} To verify the effectiveness of our designs, we conduct experiments on four datasets using several well-established HDR reconstruction models. These include AHDR \cite{yan2019attention}, which is representative of CNN-based methods, and HDR-Trans \cite{liu2022ghost}, which exemplifies Transformer-based methods. Additionally, we include TMRNet \cite{BracketIRE}, which is currently the SOTA method for HDR image reconstruction and enhancement. {We further evaluate our method against a rich collection of SOTA techniques \cite{yan2020deep,10288540,ni2025semantic,hu2024generating,li2025afunet,yan2023unified,Tel_2023_ICCV,HDRGAN,yan2020deep,Kalantari2017Deep}.  
Specifically, for Zhang \etal’s dataset \cite{BracketIRE}, we additionally include two models: Burstformer \cite{dudhane2023burstormer} and BIPNet \cite{dudhane2022burst}, as representative works in burst image restoration and enhancement. We also incorporate Kim \etal’s method \cite{kim2023joint}, which is specifically designed for RAW-domain HDR tasks (\eg, demosaicing and deghosting) and thus aligns with the dataset’s RAW format.
These models and datasets are carefully selected to cover a range of HDR reconstruction methodologies and scenarios, thereby demonstrating the versatility of our framework.}

\begin{table}[!t]
\centering
\caption{The evaluation results on Zhang's dataset.}
\scalebox{1}{
\setlength{\tabcolsep}{1.2mm}
\begin{tabular}{ccc}
\toprule[1.2pt]
\multirow{2}{*}{Methods}    & BracketIRE \cite{BracketIRE}            & BracketIRE+\cite{BracketIRE}           \\ \cline{2-3} 
                            & PSNR$\uparrow$/SSIM$\uparrow$/LPIPS$\downarrow$& PSNR$\uparrow$/SSIM$\uparrow$/LPIPS$\downarrow$\\ \hline
HDRGAN \cite{niu2021hdr}                      & 35.05 / 0.9161 / 0.175    & 28.35 / 0.8423 / 0.338    \\
BIPNet \cite{dudhane2022burst}                     & 35.97 / 0.9314 / 0.145    & 28.44 / 0.8453 / 0.311    \\
Burstormer \cite{dudhane2023burstormer}                 & 37.01 / 0.9454 / 0.127    & 28.59 / 0.8516 / 0.292    \\
SCTNet \cite{Tel_2023_ICCV}                     & 36.90 / 0.9437 / 0.120    & 28.28 / 0.8461 / 0.282    \\
Kim \etal \cite{kim2023joint}                        & 37.93 / 0.9452 / 0.115    & 28.33 / 0.8494 / 0.270    \\ \midrule \midrule
AHDR \cite{yan2019attention}                       & 36.32 / 0.9273 / 0.154    & 28.17 / 0.8422 / 0.309    \\ \hline
\multirow{2}{*}{AHDR+}      & 38.78 / 0.9480 / 0.116    & 30.25 / 0.8696 / 0.274    \\
                            & \red{+2.46} / \red{+2.07} / \red{-0.038}     & \red{+2.08} / \red{+2.74} / \red{-0.035}     \\ \midrule \midrule
HDR-Trans \cite{liu2022ghost}                  & 36.54 / 0.9341 / 0.127    & 28.18 / 0.8483 / 0.282    \\ \hline
\multirow{2}{*}{HDR-Trans+} & 39.09 / 0.9507 / 0.104    & 30.34 / 0.8706 / 0.270    \\
                            & \red{+2.55} / \red{+1.66} / \red{-0.023}     & \red{+2.16} / \red{+2.23} / \red{-0.012}     \\ \midrule \midrule
TMRNet \cite{BracketIRE}                     & 38.19 / 0.9488 / 0.112    & 28.91 / 0.8572 / 0.273    \\ \hline
\multirow{2}{*}{TMRNet+}    & 39.71 / 0.9527 / 0.110    & 30.61 / 0.8723 / 0.269    \\
                            & \red{+1.42} / \red{+0.39} / \red{-0.002}     & \red{+1.70} / \red{+1.41} / \red{-0.004}    \\
                            
\toprule[1.2pt]
\end{tabular}}
\label{tabelcompare2}
\end{table}

\noindent\textbf{Implementation Details.}
We train all components of our method simultaneously, utilizing the released code for the baseline networks with consistent training settings. We obtain semantic intermediate features and semantic maps through the frozen, pre-trained FastSAM \cite{zhao2023fastsegment}. The training is conducted using PyTorch in a Python environment, leveraging four NVIDIA Tesla A100 GPUs. We employ the Adam optimizer, configured with $\beta_1 = 0.9$ and $\beta_2 = 0.999$, for optimization. A learning rate of $2 \times 10^{-4}$ is applied, and each GPU processes a batch size of 16. Notably, the BracketlRE+ dataset from Zhang \etal requires an additional super-resolution step. For all baseline models, the $Upsampler$ module is implemented using the default method provided in the benchmark \cite{BracketIRE}. 

\subsection{Quantitative Evaluation}
\noindent\textbf{Quantitative results on RAW format dataset.} The evaluation results on Zhang \etal's dataset are shown in Tab. \ref{tabelcompare2}. The benchmark matches the results of the NTIRE \cite{zhang2024ntire} version and includes official weights.
By introducing the semantic prior of self-distillation learning at the content and color levels, our method provides an average improvement of 2.14/1.98 dB in PSNR on the BracketIRE/BracketIRE+ datasets, respectively. Furthermore, TMRNet+ achieves PSNR values of 39.71/30.61 dB on the BracketIRE/BracketIRE+ datasets, setting a new state-of-the-art benchmark. Furthermore, our framework significantly enhances the SSIM index, underscoring its superior ability to restore the image dynamic range while maintaining structural integrity and fine details. Additionally, the notable increase in LPIPS scores, facilitated by our framework, demonstrates a closer alignment with human perceptual judgment. This is achieved through the innovative incorporation of semantic priors into our design.

\noindent\textbf{Quantitative results on sRGB format dataset.}
{The evaluation results on Tel \etal’s and Kalantari \etal’s datasets are presented in Tab. \ref{compare_1} and Tab. \ref{compare_2}, respectively. It can be observed that our framework effectively enhances the performance of baseline methods, achieving an average improvement of 1.26 dB and 0.92 dB in PSNR-PU on Kalantari \etal’s and Tel \etal’s datasets, respectively.
Notably, when applied to our proposed framework, TMRNet yields a relatively smaller performance gain compared to AHDR and HDR-Transformer. This phenomenon can be attributed to the fact that TMRNet is inherently designed for HDR reconstruction scenarios with degradations, whereas the latter two baseline models focus more on dynamic motion scenes. Consequently, the performance improvement brought by semantic prior injection is relatively limited for TMRNet.
Nevertheless, our proposed framework consistently enhances the performance of all baseline models. By delivering consistent and stable performance improvements across multiple scene-specific datasets, our framework fully demonstrates its effectiveness and flexibility.}
\begin{figure*}[t]
\centering
\includegraphics[width=1\textwidth]{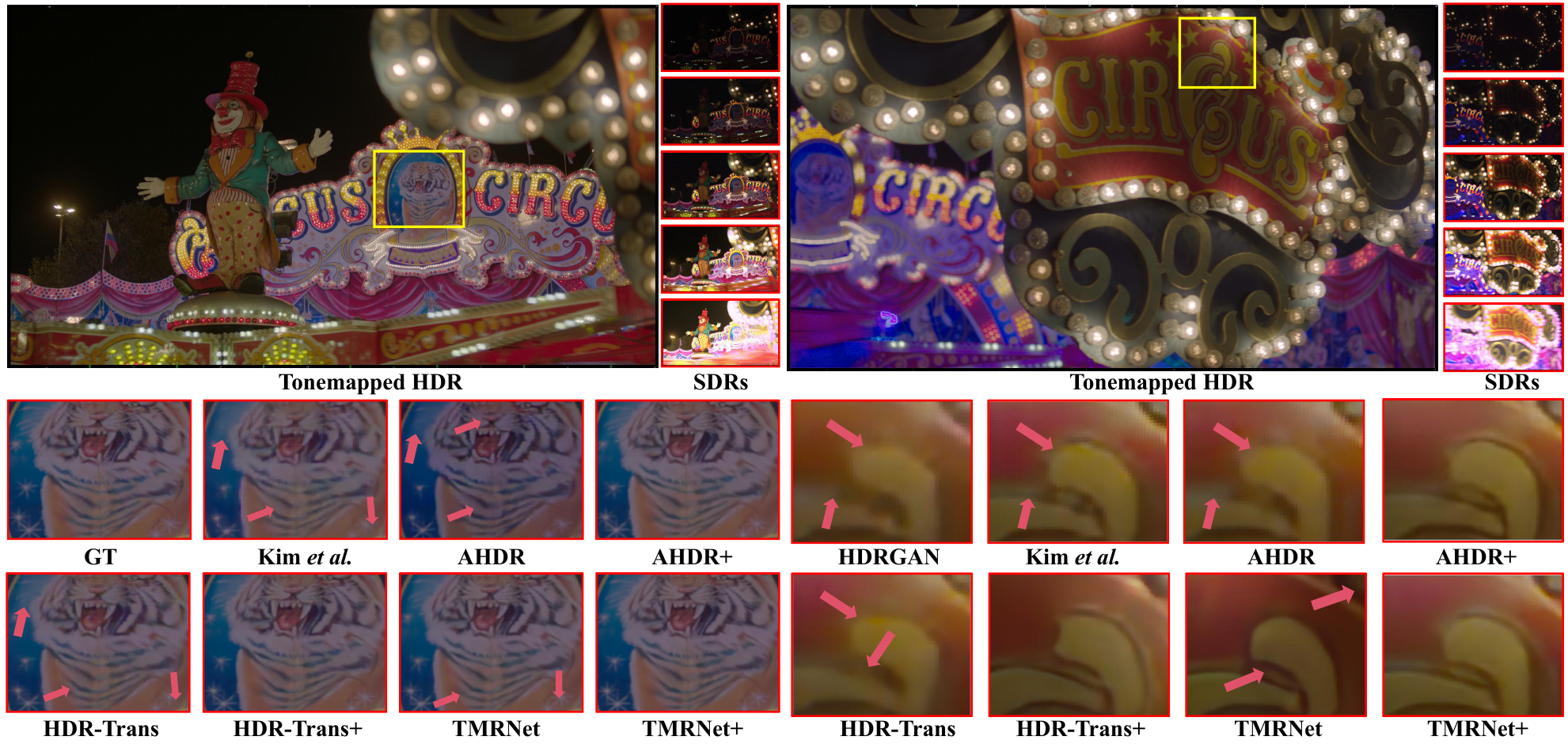} 
\caption{{Visual comparison of baseline methods with and without our framework on BracketIRE and BracketIRE+ datasets.}}
\label{compareB}
\end{figure*}

\subsection{Qualitative Evaluation}
\noindent\textbf{Qualitative results on bright scenes.}
Reconstructing HDR images in dynamic bright scenes often causes ghosting artifacts when essential content (\eg, content, details) for overexposed areas is missing due to object or camera movement. Fig. \ref{compareA} showcases qualitative results for bright scenes from the two datasets. Our framework is applied to multiple baseline models and compared with various state-of-the-art approaches using test data from the Kalantari \etal’s  (top row) and Tel \etal’s (bottom row) datasets, which include challenging samples characterized by saturated backgrounds and foreground motions.
As shown in the top row of Fig. \ref{compareA}, all baseline results exhibit noticeable artifacts in the step regions, whereas our proposed framework significantly suppresses ghosting phenomena. In the bottom row, all improved methods except AHDR+ eliminate ghosting, but AHDR+ still shows a significant reduction in ghosting compared to the baselines. This demonstrates the effectiveness of our framework in removing or suppressing ghosting caused by missing information.

\noindent\textbf{Qualitative results on low-light scenes.}
Captured SDR images often suffer from noticeable degradation in low-light scenes, which significantly affects the final reconstruction quality. Fig. \ref{compareB} showcases qualitative results from the BracketIRE (left) and BracketIRE+ (right) datasets. As shown in the left part, the SDR images contain noticeable degradations, and all methods exhibit unclear details. Our proposed framework significantly enhances the model's detail representation, reducing the impact of image degradations. For example, the tigers and stars within the blue boxes exhibit clearer textures.
Furthermore, compared to BracketIRE, the SDR images in BracketIRE+ have lower resolution, posing a challenge for reconstructing high-quality HDR images. As shown in the right part, all baseline models almost fail to restore realistic textures and show significant blurring. In contrast, our proposed framework significantly improves the imaging quality of the baseline models, recovering the scene's outlines and details more completely.

\begin{figure*}[t]
\centering
\includegraphics[width=\textwidth]{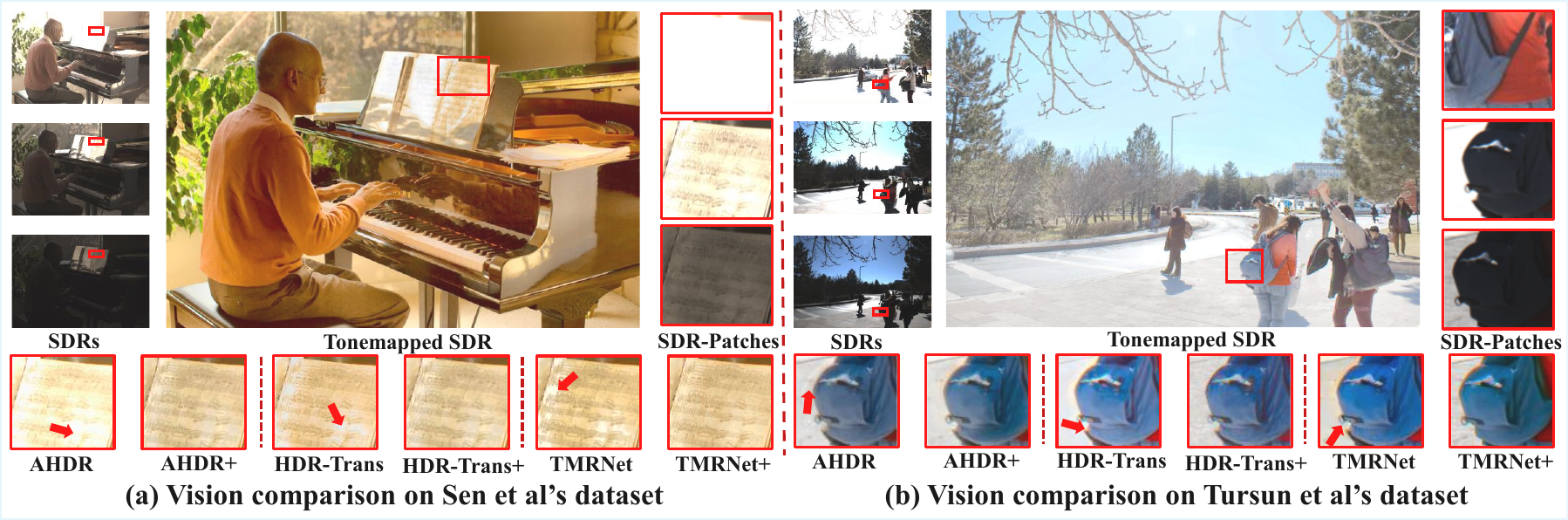}
\caption{{Qualitative generalization results on Sen \etal's dataset\cite{sen2012robust} and Tursun \etal's dataset\cite{Tursun2016data}.}}
\label{fig:generalization_realworld}
\vspace{+0.1cm}
\end{figure*}

{\noindent\textbf{Qualitative results on dataset without ground truth.}
To further validate the generalization capability of the proposed method, we additionally assess its performance using the datasets from Sen \etal \cite{sen2012robust} and Tursun \etal \cite{Tursun2016data}, neither of which provides ground truth, as shown in Fig.~\ref{fig:generalization_realworld}. On the Sen \etal' dataset (left), severe overexposure and fine-scale textures cause AHDR~\cite{yan2019attention}, HDR-Trans~\cite{liu2022ghost}, and TMRNet~\cite{BracketIRE} to produce blown-out regions with texture loss or over-exposure. In contrast, when our framework is applied to distill semantic priors from FastSAM~\cite{zhao2023fastsegment} and enforce color-consistency regularization, these models recover clearer details, stabilize local contrast, and yield more natural highlight roll-off.
On the Tursun \etal's dataset (right), extreme luminance gradients cause AHDR and TMRNet to generate ghosting near the backpack boundary, while HDR-Trans loses fine structure. After integrating our framework, the same baselines preserve sharp boundaries, maintain structural coherence, suppress ghosting, and reconstruct smooth and physically plausible luminance gradients.
These results demonstrate that our framework generalizes well to challenging real-world scenes with complex exposure conditions and motion, consistently improving both visual quality and structural fidelity.}

\begin{table}[t]
\centering
\caption{{Ablation studies of proposed framework.}}
\label{ab}
\scalebox{0.85}{
\begin{tabular}{ccc|cc}

\toprule[1.2pt]
\multirow{2}{*}{Method} & \multicolumn{2}{c|}{ORM Results} & \multicolumn{2}{c}{SPGRM Results} \\ \cline{2-5} 
                        & PSNR-PU $\uparrow$ & SSIM-PU $\uparrow$ & PSNR-PU $\uparrow$& SSIM-PU $\uparrow$       \\ \hline
AHDR\cite{yan2019attention}                & 42.29      & 0.9919     & -         & -          \\
+ Feature Injection & 42.45      & 0.9921     & -         & -          \\
+ Segmentation Map & 42.39      & 0.9923     & -         & -          \\ \hline
w/o $\mathcal{L}_{\text {content}}$                  & 42.46      & 0.9920     & 42.98     & 0.9925     \\
w/o $\mathcal{L}_{\text {color}}$               & 43.41      & 0.9926     & 43.59     & 0.9928     \\
w/o SKAM                & 43.53      & 0.9928     & 43.57     & 0.9931     \\
Direct Alignment        & 39.69      & 0.9890     & 41.63     & 0.9907     \\
w/o FastSAM\cite{zhao2023fastsegment}                 & 42.23      & 0.9915     & 42.21     & 0.9913     \\
w HRNet\cite{wang2020deep}                 & 43.29      & 0.9929     & 43.28     & 0.9929     \\ \hline
AHDR+                   & \textbf{43.61}      & \textbf{0.9937}     & \textbf{43.62}     & \textbf{0.9937}     \\ \toprule[1.2pt]
\end{tabular}}
\end{table}

\subsection{Ablation Studies}
Considering that Tel's dataset offers a more balanced distribution of scenes and motions,
We conduct ablation experiments on this dataset to validate the effectiveness of our proposed method. 

\noindent\textbf{Effects of Semantic Knowledge Transfer Scheme.}
To bridge the performance gap between the teacher and the student model, we introduce two loss functions: $\mathcal{L}_{\text{color}}$ and $\mathcal{L}_{\text{content}}$. These losses facilitate semantic knowledge transfer at the color and content levels, respectively. To assess their impact on model performance, we conduct experiments by selectively removing these loss functions and analyzing the results.
As presented in Tab \ref{ab}, $\mathcal{L}_{\text{content}}$, which is a widely adopted approach in existing distillation schemes, serves as the foundational loss function for self-distillation. It significantly influences the model's overall performance. Meanwhile, while relying solely on $\mathcal{L}_{\text{color}}$ does not suffice for efficient self-distillation, combining it with $\mathcal{L}_{\text{content}}$ yields a noticeable performance boost. This suggests that $\mathcal{L}_{\text{color}}$ effectively addresses the challenge of dramatic brightness variations across different regions of HDR images, enhancing the model's ability to capture color consistency and detail in challenging lighting conditions.

\noindent\textbf{Effects of Semantic Knowledge Alignment Module.}
In addition to model-level distillation, the Semantic Knowledge Alignment Module further enhances self-distillation by mapping features from different domains into a shared latent space for semantic alignment. As shown in Tab. \ref{ab}, the variant labeled ``W/o SKAM" refers to the model with the SKAM removed. We observe that this variation slightly degrades the performance of the ORM, suggesting that SKAM plays a key role in facilitating effective semantic knowledge transfer.
Furthermore, we assessed the impact of directly aligning the features of different models using Mean Squared Error loss without SKAM, denoted as ``Direct Alignment." The results indicate that, due to the domain gap between the intermediate features of different models, direct alignment significantly impairs the performance of both the student and teacher models. In contrast, SKAM employs an encoder to map $F_S^k$ and $F_T^k$ to a shared latent space. A small loss in this setup demonstrates that $F_S^k$ can effectively reconstruct the masked parts of $F_T^k$ within the latent space, thereby achieving semantic alignment between $F_T^k$ and $F_S^k$.

\noindent\textbf{Effects of Semantic Knowledge Bank.}
The proposed framework aims to address the ill-posed nature of High Dynamic Range reconstruction by leveraging semantic knowledge from the Semantic Knowledge Bank. To validate its effectiveness, we performed ablation studies by removing the semantic priors and extracting the semantic priors from different models. As shown in Tab.\ref{ab}, ``w/o FastSAM" indicates the removal of semantic priors in the SPGRM, instead employing a simple second-stage enhancement strategy. The results demonstrate that merely increasing the model's parameter count does not lead to substantial performance gains and may even cause overfitting, resulting in performance degradation.
In addition, we evaluated the influence of various semantic priors derived from FastSAM and the widely used HRNet \cite{wang2020deep}. Given that the SAM-based method provides richer and more nuanced semantic information across multiple scales, our approach outperforms HRNet, which relies solely on semantic cues from instance and category labels.

{\noindent\textbf{Effects of Semantic Prior Injection Strategy.}
We conducted additional ablation experiments on the  kalantari \etal's\cite{kalantari2019deep} dataset using the AHDR\cite{yan2019attention} baseline. Specifically, we integrated semantic information in two standard ways common in low-level vision. First is semantic feature injection: we extracted semantic features from each input SDR frame via a pre-trained FastSAM\cite{zhao2023fastsegment} encoder, fused them using an FPN, and used the resulting representation to modulate AHDR's intermediate features through affine transformations. Second is semantic segmentation map guidance: we generated per-SDR segmentation masks with FastSAM and concatenated them with SDR inputs as additional channels to guide HDR reconstruction. As shown in Table \ref{ab}, both variants yield a modest improvement in PSNR-PU but still lag significantly behind our proposed framework. We attribute this to two key challenges in multi-exposure HDR reconstruction: (1) Input SDR images suffer from exposure-related degradations, such as under/over-exposure and noise, which makes direct extraction of reliable semantic priors difficult. (2) Frame misalignment, caused by camera or object motion, leads to inconsistencies in per-frame semantic estimates and conflicting guidance during fusion. In contrast, our method extracts semantic priors from a static, initial HDR estimate. This estimate integrates information from all exposures, yielding a more robust and coherent semantic representation that is less affected by degradations or motion.}
\begin{figure*}[t]
\centering
\includegraphics[width=1\linewidth]{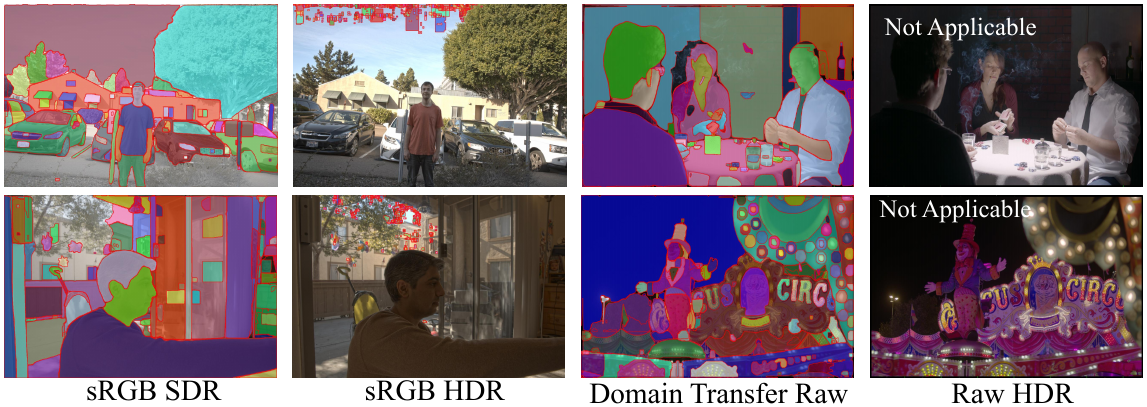} 
\caption{Segmentation results of HDR images from different domains and formats in FastSAM.}
\label{tone}
\end{figure*}

\subsection{Discussion}
\noindent\textbf{Computational Overhead of Training Stage.}
In the proposed framework, we utilize semantic knowledge extracted from a pre-trained semantic model to tackle the ill-posed problem of HDR reconstruction. Upon completion of training, our approach substantially improves imaging quality without necessitating alterations to the original model architecture or introducing additional computational overhead during inference. However, the training phase incurs extra computational costs due to the need to train the Semantic Prior Guided Reconstruction Module and conduct inference using FastSAM.
We detail the training computational overhead in Table \ref{tab:overhead}, based on an NVIDIA A100 GPU with 128×128 image patches and a batch size of 8. The primary source of this overhead is the SPGRM, which requires preprocessing of semantic prior features and reconstructs high-quality HDR images under their guidance. In comparison, the Semantic Knowledge Alignment Module imposes a relatively minor computational burden, as it conducts feature semantic alignment in the latent space using only a small number of convolutional layers. Additionally, while FastSAM possesses a significantly larger parameter count than the original model (68M \textit{vs.} 1.24M), its weights are frozen during training, eliminating the need for gradient computations and restricting its role to inference. Consequently, FastSAM contributes minimal computational overhead and negligible memory usage.
{When using AHDR as the baseline model, the total time to complete the entire training process on a single NVIDIA A100 GPU is approximately 20 hours. This total duration varies depending on the chosen baseline model. For instance, if the baseline model is inherently computationally intensive (\eg, HDR-Transformer \cite{liu2022ghost}), the relative proportion of the additional overhead introduced by our framework in the total time will be comparatively smaller.}

Overall, although our framework introduces increased computational demands and memory consumption during training stage, it offers a notable advantage in resource efficiency compared to retraining a semantic prior model tailored specifically for HDR tasks. Furthermore, in the field of HDR reconstruction, existing methods under the same paradigm typically improve model performance at the cost of additional inference overhead. In contrast, our approach achieves comparable improvements without modifying the network architecture, thereby avoiding any extra computational overhead during inference.

\noindent\textbf{Choice of Domain Transfer Operation.}
Due to the inherent differences in domain and format between HDR and SDR images, our framework requires preprocessing HDR images to obtain priors and aligning the outputs of different models before performing distillation. The choice of domain transfer operation for this alignment is critical to the success of our approach. For sRGB HDR images, this involves a tone mapping operation, while for RAW images, an additional demosaicing operation is included.
 
In our method, we adopt the $\mu$-law for tone mapping due to its proven effectiveness and simplicity. The $\mu$-law function has been widely utilized in prior work \cite{yan2019attention,Kalantari2017Deep,HDRGAN} for loss function computations, demonstrating its robustness. Although more advanced tone mapping techniques exist, they are often complex and non-differentiable, making them unsuitable for our framework. For instance, Gamma encoding, despite its simplicity, is non-differentiable near zero, limiting its applicability.
Additionally, converting RAW images to sRGB requires multiple operations, such as demosaicing and white balancing. Among these, demosaicing is particularly crucial, as it directly impacts the structural integrity of the image. In contrast, white balancing relies on additional metadata and primarily affects the image's color tone without altering its content. Our experiments show that white balancing has minimal impact on segmentation results. Consequently, we incorporate only an additional demosaicing operation alongside tone mapping to process RAW images effectively. As illustrated in Fig. \ref{tone}, HDR images processed with the proposed domain transfer operation can be effectively utilized for semantic knowledge extraction, demonstrating the robustness of our method.

\noindent \xblue{\textbf{Discussion on GAN-based Baselines.} While our framework demonstrated robust improvements on CNN-based (\eg, AHDR~\cite{yan2019attention}) and Transformer-based (\eg, HDR-Trans~\cite{liu2022ghost}) models, we encountered challenges when applying it to GAN-based methods, specifically HDRGAN~\cite{niu2021hdr}, which is widely recognized as an excellent and influential method in the field of HDR reconstruction. Despite its theoretical merits, we observed severe training instability during our integration experiments. This sensitivity is likely due to the inherent difficulty in balancing the generator and discriminator during adversarial training, which requires extensive hyperparameter tuning and careful optimization scheduling. We report this negative result to highlight the potential limitations of applying our current framework to GAN-based models, suggesting that specialized stabilization techniques may be required for such architectures.}

\noindent\textbf{Limitations and Future Work}
{A key limitation of the proposed framework lies in the increased computational overhead during training: compared to the original reconstruction model, the SPGRM and semantic feature preprocessing add extra training time, though inference remains efficient by only using the optimized ORM. Another limitation is its limited support for HDR-specific wide-gamut color spaces (\eg, Rec. 2020, DCI-P3) that are widely used in professional HDR scenarios, current experiments focus on sRGB/RAW formats, and the existing $\mu$-law domain transfer cannot fully leverage the dynamic range of these advanced color spaces. For future work, we will explore more lightweight knowledge transfer strategies to reduce training overhead while preserving semantic cues, and extend the domain transfer module to support Rec. 2020/DCI-P3, further validating the framework on professional HDR datasets.}

\begin{table}[t] 
\caption{Additional computational overhead of training stage.}
\resizebox{\columnwidth}{!}{
    \begin{tabular}{c|cccc}
    \toprule
    
Module       & Para. & Flops & GPU Memory & Time \\ \midrule
    AHDR    &1.24M            &163.45G       &2.38GB            &31.8ms                \\
    SPGRM+FPN   &1.44M            &151.09G       &4.76GB            &69.5ms                \\
    SKAM   &0.67M            &87.32G       &0.88GB            &17.6ms                \\
    FastSAM &68M            &76.77G       &0.32GB            &30.19ms                \\ \bottomrule
    \end{tabular}
}
\label{tab:overhead}
\end{table}
\section{Conclusion}
Existing multi-exposure HDR imaging methods primarily focus on mitigating artifact issues caused by motion, but they overlook the inherently ill-posed nature of captured SDR images in real-world scenarios. To tackle these challenges, we introduce a general framework that leverages semantic priors from pretrained segmentation models to boost existing HDR reconstruction models without modifying the original network. By endowing models with semantic knowledge, our framework enables a deeper understanding of image content, providing explicit guidance for the reconstruction process. Extensive experiments demonstrate that the proposed framework consistently improves the performance of multiple baseline models, achieving state-of-the-art results across diverse datasets encompassing various scenes and formats.

\section*{Acknowledgment}
This work is supported by NSFC of China under Grant 62301432 and Grant 62306240, the Natural
Science Basic Research Program of Shaanxi under Grant 2023-JC-QN-0685 and Grant QCYRCXM-2023-057, the Fundamental Research
Funds for Central Universities, and Guangdong Basic and Applied Basic Research Foundation 2025A1515011119, the Innovation Foundation for Doctor Dissertation of Northwestern Polytechnical University ZX2025019.

\bibliographystyle{IEEEtran}
\bibliography{egbib.bib}

\end{document}